\documentclass{article}

    \PassOptionsToPackage{numbers, compress}{natbib}
 \usepackage[dblblindworkshop, final]{neurips_2025}

\usepackage[utf8]{inputenc} % allow utf-8 input
\usepackage[T1]{fontenc}    % use 8-bit T1 fonts
\usepackage{hyperref}       % hyperlinks
\usepackage{url}            % simple URL typesetting
\usepackage{booktabs}       % professional-quality tables
\usepackage{amsfonts}       % blackboard math symbols
\usepackage{nicefrac}       % compact symbols for 1/2, etc.
\usepackage{microtype}      % microtypography
\usepackage[dvipsnames]{xcolor}
\usepackage{booktabs}
\usepackage{graphicx}
\usepackage{multirow}
\usepackage{amsmath}
\usepackage{subcaption}
\usepackage{listings}
\usepackage{lscape}
\usepackage{longtable}
\usepackage{placeins}

\lstdefinestyle{promptstyle}{
  basicstyle=\ttfamily\small,        % monospaced, slightly smaller for fit
  breaklines=true,                    % wrap long lines
  breakatwhitespace=false,
  postbreak=\mbox{\textcolor{gray}{$\hookrightarrow$}\space},
  xleftmargin=2mm, frame=single,     % subtle frame + small left indent
  framesep=2mm,
  columns=fullflexible,               % better handling for long tokens
  tabsize=2,
  showstringspaces=false,
  keepspaces=true                     % preserve indentation
}

\title{When Agents go Astray: \\Course-Correcting SWE Agents with PRMs}

\workshoptitle{Scaling Environments for Agents (SEA)}

\author{%
  Shubham Gandhi\thanks{Work done as an intern at IBM Research.} \\
  Carnegie Mellon University \\
  \texttt{srgandhi@andrew.cmu.edu} \\
  \And
  Jason Tsay \\
  IBM Research \\
  \texttt{jason.tsay@ibm.com} \\
  \And
  Jatin Ganhotra \\
  IBM Research \\
  \texttt{jatinganhotra@us.ibm.com} \\
  \And
  Kiran Kate \\
  IBM Research \\
  \texttt{kakate@us.ibm.com} \\
  \And
  Yara Rizk \\
  IBM Research \\
  \texttt{yara.rizk@ibm.com} \\
}

\begin{document}

\maketitle

\begin{abstract}
% \jatin{Code Agents -> SWE Agents in Title?} 
\looseness=-1 Large Language Model (LLM) agents are increasingly deployed for complex, multi-step software engineering (SWE) tasks. However, their trajectories often contain costly inefficiencies, such as redundant exploration, looping, and failure to terminate once a solution is reached. Prior work has largely treated these errors in a post-hoc manner, diagnosing failures only after execution. In this paper, we introduce \textbf{\texttt{SWE-PRM}}, an inference-time Process Reward Model (PRM) that intervenes during execution to detect and course-correct trajectory-level errors. Our PRM design leverages a taxonomy of common inefficiencies and delivers lightweight, interpretable feedback without modifying the underlying policy. On SWE-bench Verified, closed-source PRMs improve resolution from 40.0\% to 50.6\% (+10.6 p.p.), with the largest gains on medium and hard tasks. Among feedback strategies, taxonomy-guided PRMs outperform unguided or explicit action-prescriptive variants, increasing success rate while reducing trajectory length. These benefits come at an acceptable added inference cost of as low as \$0.2, making PRMs a practical and scalable mechanism for improving SWE agents' reliability and efficiency.

\end{abstract}

\section{Introduction}

Large Language Model (LLM)-based agents are increasingly deployed for complex, multi-step software engineering (SWE) tasks, such as repository-level bug fixing and feature implementation \cite{jimenez2024swebench, zan2025multiswebenchmultilingualbenchmarkissue, rashid2025swepolybenchmultilanguagebenchmarkrepository, li2025feabenchbenchmarkevaluatingrepositorylevel, gautam2025refactorbenchevaluatingstatefulreasoning, deng2025nocodebenchbenchmarkevaluatingnatural}. While recent advances have improved benchmark resolution rates, these gains often mask hidden inefficiencies in the agent’s execution process. In particular, \emph{trajectory-level errors}, i.e. patterns such as action looping, redundant backtracking, or drifting toward irrelevant subgoals, can accumulate over a run. On top of yielding incorrect actions, these behaviors also waste compute, inflate latency, and risk exhausting the agent’s budget before task completion.

Prior work on SWE agents has largely focused on maximizing \emph{success rate} without explicitly addressing process efficiency. For example, systems such as SWE-smith \cite{yang2025swesmithscalingdatasoftware}, SWE-gym \cite{pan2025training}, and R2E-gym \cite{jain2025r2e} train an open source model to reduce inference cost, but high success rates do not guarantee low-cost, efficient execution. This gap is particularly significant because trajectory-level inefficiencies have been documented for SWE tasks \cite{deshpande2025trailtracereasoningagentic} and noted in other sequential decision-making domains \cite{cemri2025multiagentllmsystemsfail}, suggesting that a mitigation strategy like ours could generalize beyond SWE.

Existing approaches for handling trajectory-level errors focus on \emph{post-mortem} analysis. For example, TRAIL \cite{deshpande2025trailtracereasoningagentic} and MAST \cite{cemri2025multiagentllmsystemsfail} rely on dumping the entire trajectory to an LLM judge for error analysis after execution. While useful for research diagnostics, these methods are impractical in deployment: they incur substantial context-length overhead, require expensive iterative re-judging, and cannot prevent wasted computation that has already occurred. In practice, the iterative cycle often involves a human analyst reviewing error reports and manually adjusting prompts, heuristics, or control logic between runs. This is fundamentally different from our setting, where the base agent remains fixed during execution, and intervention is applied only through lightweight, inference-time guidance.

Other strategies for guiding agent behaviour also have limitations. \emph{Outcome Reward Models} (ORMs) focus solely on evaluating final solutions for correctness, ignoring process optimality and therefore missing costly but non-terminal inefficiencies \cite{lightman2023let}. Some methods use \emph{Process Reward Models} (PRM) within Monte-Carlo Tree Search (MCTS) to score multiple future rollouts per step \cite{antoniades2024swe}; however, for SWE agents this is prohibitively expensive. Code-editing actions are often irreversible, making it infeasible to spin up parallel environment instances or reset to arbitrary intermediate states without high overhead.

In this work, we propose an \emph{inference-time PRM}, \texttt{SWE-PRM}, that \textbf{prevents}, \textbf{detects}, and \textbf{course-corrects} trajectory-level errors \emph{during} execution. The PRM is invoked periodically with a limited sliding window of past steps and is guided by a taxonomy of common error patterns. It issues actionable feedback that can be applied immediately, steering the agent back toward efficient completion without modifying its core architecture or parameters. To the best of our knowledge, this is the \emph{first} application of PRMs for real-time trajectory-level error correction in SWE agents. Our design offers three advantages: (1) \textbf{real-time mitigation} of errors before they propagate, (2) \textbf{cost-efficiency} through sparse, targeted PRM calls, and (3) \textbf{modularity} for integration with both open-weight and proprietary LLMs, making it potentially transferable to other domains where similar inefficiencies have been observed.

\looseness=-1 We evaluate \texttt{SWE-PRM} on the SWE-bench Verified benchmark using \textsc{SWE-agent-LM-32B}, a finetuned \textsc{Qwen2.5-Coder-32B-Instruct} model as the policy model \cite{yang2025swesmithscalingdatasoftware}. We compare open-weight and frontier models as PRMs, with and without taxonomy guidance. Our results show that a strong PRM significantly improves resolution rate and cost-effectiveness over both a base SWE-agent and post-hoc analysis baselines, with consistent gains across all categories: easy, medium, and hard instances. Concretely, our experiments show that a taxonomy-guided PRM improves resolution from 40.0\% to 50.6\% on SWE-bench Verified, including +10.7 points on medium and +4.4 points on hard tasks. These gains come with shorter or comparable trajectories, translating into more efficient runs. While PRM guidance adds inference cost, the additional spend amounts to roughly \$0.2 per instance, highlighting PRMs as an attractive tradeoff between accuracy and efficiency in long-horizon SWE agents.

\begin{figure}
    \centering
    \includegraphics[width=1\linewidth]{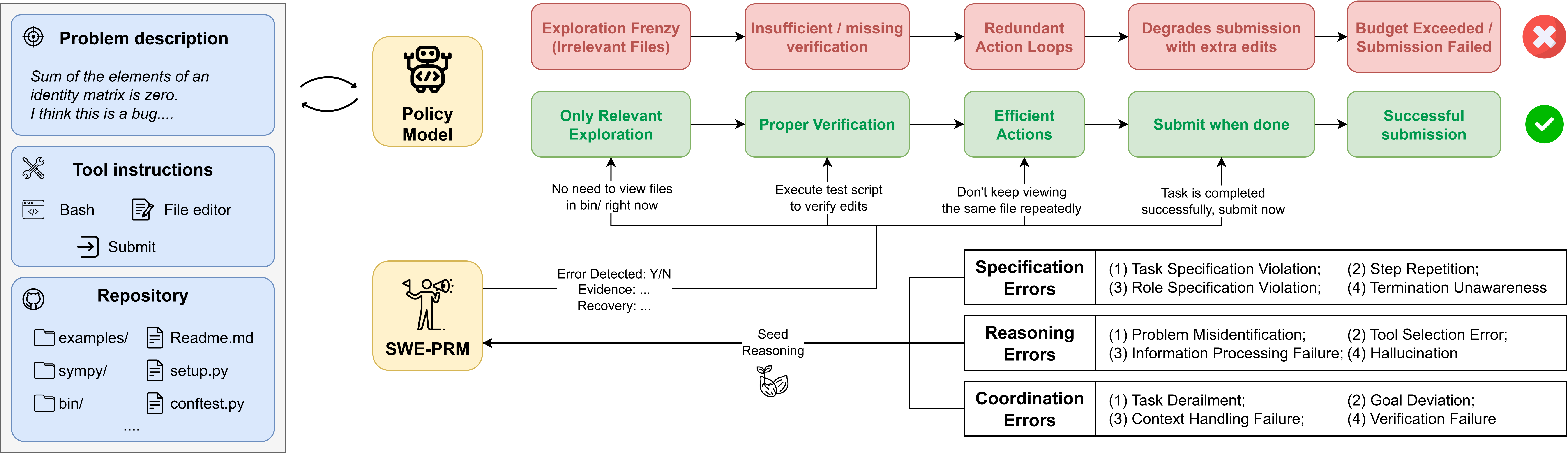}
    \caption{\texttt{SWE-PRM} helps mitigate trajectory-level suboptimalities in SWE agents.}
    \label{fig:prms-for-swe}
\end{figure}
\section{Related Work}

\subsection{Repository-Level Code Generation}

Repository-level software engineering benchmarks have driven much of the recent progress in code agents. SWE-bench \cite{jimenez2024swebench} provides realistic bug-fixing and feature implementation tasks from open-source repositories, with deterministic evaluation for correctness. Since SWE-bench, several new benchmarks have emerged to broaden repository-level evaluation: Multi-SWE-bench \cite{zan2025multiswebenchmultilingualbenchmarkissue} extends issue-resolving tasks to multiple programming languages, SWE-PolyBench \cite{rashid2025swepolybenchmultilanguagebenchmarkrepository} introduces multi-language tasks with syntax tree analysis-based metrics, FEA-Bench \cite{li2025feabenchbenchmarkevaluatingrepositorylevel} focuses on repository-level feature implementation, RefactorBench \cite{gautam2025refactorbenchevaluatingstatefulreasoning} targets multi-file refactoring, and NoCode-bench \cite{deng2025nocodebenchbenchmarkevaluatingnatural} evaluates natural language-driven feature addition. SWE-gym \cite{pan2025training} offers a training and evaluation framework for coding agents and verifiers, while R2E-gym \cite{jain2025r2e} introduces procedural environments with hybrid verifiers to facilitate scaling open-weight agents. However, these benchmarks and frameworks primarily aim to improve final resolution rates and do not directly address execution efficiency or trajectory-level inefficiencies, which is the focus of our approach. In an effort to replace frontier models and achieve good performance with open-source models, SWE-Smith \cite{yang2025swesmithscalingdatasoftware} scales data generation for code agents and releases \texttt{SWE-agent-LM-32B}, the policy model we use in our experiments (a finetuned version of \texttt{Qwen2.5-Coder-32B-Instruct}).

\subsection{Improving LLM Agents}
A number of works have sought to improve the performance, robustness, and reasoning quality of LLM agents.  

\paragraph{Error analysis and taxonomies.}  
\citet{deshpande2025trailtracereasoningagentic} introduces a comprehensive taxonomy of reasoning, execution, and planning errors in SWE agents, with human-annotated traces from SWE-bench and GAIA. \citet{cemri2025multiagentllmsystemsfail} proposes a taxonomy for multi-agent LLM systems, emphasizing coordination and reasoning failures. Both rely on post-mortem trajectory dumps to an LLM judge, often combined with human review, which limits their ability to prevent wasted computation during execution. \citet{chen2025unveilingpitfallsunderstandingaidriven} similarly analyses common failure modes of code agents on real-world GitHub issues, while \citet{sung2025verilahumancenteredevaluationframework} proposes VeriLA, a human-aligned verification framework for making agent failures more interpretable. These works highlight the need for systematic, taxonomy-guided diagnostics, but remain primarily retrospective.

\paragraph{Search-based improvements.}
\citet{antoniades2025swesearchenhancingsoftwareagents} integrate Monte Carlo Tree Search (MCTS) with self-assessment to explore multiple candidate solution paths in SWE agents, yielding substantial performance gains without additional model training. \citet{zainullina2025guidedsearchstrategiesnonserializable} address search in non-serializable environments by introducing one-step lookahead and trajectory selection policies guided by learned action-value estimators, achieving improved results on SWE-bench Verified. While effective, these methods can be costly for long-horizon, irreversible settings such as repository-level code editing.

\paragraph{Process optimization and recovery.}
BacktrackAgent \cite{wu2025backtrackagentenhancingguiagent} introduces explicit verification, judgment, and reflection mechanisms to detect errors and revert to earlier states in GUI agents. \citet{song2024trialerrorexplorationbasedtrajectory} propose exploration-based trajectory optimization that learns from failed attempts to avoid repeating mistakes. SMART \cite{qian2025smartselfawareagenttool} targets tool overuse mitigation by training agents to balance tool calls with internal reasoning, reducing unnecessary invocations while maintaining or improving performance. These approaches demonstrate the value of inference-time self-correction, though often in domains other than repository-level SWE.

\paragraph{Reward models for agent improvement.}  
Reward modeling has been used to guide agents toward better intermediate decisions across various domains. Outcome Reward Models (ORMs) prioritize final outcome correctness in a task’s result—for example, ensuring a patched program passes all tests in repository-level bug fixing \cite{ma2025thinkinglongerlargerenhancing, pan2025training}. In contrast, Process Reward Models (PRMs) evaluate each intermediate step’s quality in multi-step reasoning tasks, offering finer-grained feedback signals \cite{khalifa2025processrewardmodelsthink, sun2025freeprmtrainingprocessreward, li-etal-2025-codeprm}. CodePRM \cite{li-etal-2025-codeprm} integrates execution feedback into step-level thought scoring for single-turn code generation, improving correctness when paired with a generate-verify-refine loop. FreePRM \cite{sun2025freeprmtrainingprocessreward} trains PRMs without step-level labels, using pseudo-rewards inferred from final outcomes. STeCa \cite{wang2025stecastepleveltrajectorycalibration} calibrates trajectories at the step level by replacing suboptimal actions with improved alternatives via LLM self-reflection. ThinkPRM \cite{khalifa2025processrewardmodelsthink} augments PRMs with their own reasoning chains, outperforming discriminative baselines with far less data.

While PRMs have been embedded into expensive search procedures such as MCTS, such integration is computationally prohibitive for SWE agents due to costly environment resets. To the best of our knowledge, our work is the first to apply a PRM for \emph{real-time} trajectory-level error \emph{prevention, detection, and course-correction} in SWE agents, using taxonomy-guided, inference-time feedback without modifying the base policy model.

\section{Methodology}
\label{sec:method}

\subsection{Task and Architecture}
\label{sec:task-arch}
We study repository-level issue resolution~\citep{jimenez2024swebench}: 
given a natural language problem description $d$, a set of tool instructions $i$, 
and a snapshot of a repository $\mathcal{R}$, 
the agent must propose a patch $\hat{p}$ that satisfies the repository’s test suite $\mathcal{S}$. 
The suite contains two subsets: 
$\mathcal{S}_{pp}$ (\emph{pass-to-pass}) tests that must remain successful to preserve existing functionality, 
and $\mathcal{S}_{fp}$ (\emph{fail-to-pass}) tests that must transition from failing to passing to confirm the requested change. 
A patch $\hat{p}$ is accepted iff
\[
\forall \,\sigma \in \mathcal{S}_{pp}, \;\;\sigma(\hat{p}(\mathcal{R})) = \texttt{pass}
\quad\text{and}\quad
\forall \,\sigma \in \mathcal{S}_{fp}, \;\;\sigma(\hat{p}(\mathcal{R})) = \texttt{pass}.
\]

The base agent follows the SWE-agent framework~\citep{10.5555/3737916.3739517}, running a ReAct-style loop~\citep{yao2023react} that records an explicit transcript of reasoning and interactions. At step $t$, the transcript is
\[
\mathcal{H}_t \;=\; \big(u_1,a_1,o_1,\,u_2,a_2,o_2,\,\ldots,\,u_t,a_t,o_t\big),
\]
where $u_i$ are the model’s \textit{thoughts} (free-form reasoning), $a_i$ are \textit{actions} (tool calls), and $o_i$ are the resulting \textit{observations} (e.g., file contents, diffs, or execution outputs). The policy $\pi_\theta$ conditions on $\mathcal{H}_t$ to generate the next thought and action, $(u_{t+1}, a_{t+1}) \sim \pi_\theta(\cdot \mid \mathcal{H}_t, d, i)$. Executing $a_{t+1}$ yields $o_{t+1}$, which is appended back to the transcript. This process is strictly sequential and continues until the agent submits a patch or reaches its step budget.

The action space is designed to simulate repository-level software engineering. The agent can (i) execute shell commands with \texttt{bash}, (ii) view or edit files through a persistent \texttt{str\_replace\_editor} that supports browsing paths, inserting or replacing code, creating new files, and undoing edits, and (iii) finalize its work with a \texttt{submit} action. Upon submission, the patch is evaluated in a fresh, isolated environment.

\subsection{PRM as Course-Corrector}
\label{sec:prm-course-corrector}

Process Reward Models (PRMs) are introduced as lightweight \emph{course-correctors} within the agent’s reasoning loop. Rather than replacing the base policy or dictating procedural changes, the PRM interjects periodically with natural language guidance aimed at steering the trajectory towards the next optimal action. This guidance is (1) in natural-language with demarcated sections based on taxonomy, and (2) grounded in the current context $H_t$, for the policy model to incorporate into its own reasoning.

\subsubsection{Motivation and Taxonomy}
\label{sec:taxonomy}
Long-horizon software engineering agents frequently accumulate \emph{trajectory-level inefficiencies}, patterns of reasoning and action that may not yield immediate incorrectness but gradually erode efficiency and task success. Prior work such as Trail~\cite{deshpande2025trailtracereasoningagentic} and MAST~\cite{cemri2025multiagentllmsystemsfail} introduced taxonomies of such inefficiencies, but mainly as \emph{post-mortem analysis tools}, applied after execution to explain failure. In contrast, we operationalize inefficiency categories \emph{during execution}, enabling a Process Reward Model (PRM) to deliver corrective natural language guidance in real time. This distinction is especially crucial in repository-level code editing on SWE-bench~\cite{jimenez2024swebench}, where agents such as SWE-agent~\cite{10.5555/3737916.3739517} often require dozens of dependent steps and small inefficiencies can compound into wasted effort or cascading failures.

\looseness=-1 The taxonomy itself is domain-agnostic, reflecting common patterns of inefficiency that arise in long-horizon agentic reasoning. We validate it in the SWE setting since it provides a natural stress test, but the categories are broadly applicable across other domains where agents plan, reason, and act over extended horizons. The taxonomy was seeded in manual inspection of execution traces and emphasizes not only the \emph{failure mode} but also a corresponding \emph{recovery action}. It is organized into three families:

\paragraph{Specification Errors (violations of task setup).}  
\looseness=-1 \textit{Task specification violations} (ignoring explicit requirements), \textit{role specification violations} (acting outside intended scope), \textit{step repetition} (re-executing completed actions), and \textit{termination unawareness} (continuing after completion criteria are met).

\paragraph{Reasoning Errors (decision-making failures).}  
\textit{Problem misidentification} (misunderstanding the subtask), \textit{tool selection errors} (choosing inappropriate tools), \textit{hallucinations} (fabricating results), and \textit{information processing failures} (retrieving or interpreting evidence incorrectly).

\paragraph{Coordination Errors (multi-step process management failures).}  
\textit{Task derailment} (macro-level drift, abandoning the main task), \textit{goal deviation} (micro-level misalignment, pursuing secondary or irrelevant subgoals), \textit{context handling failures} (forgetting prior results), and \textit{verification failures} (neglecting to check correctness or quality).

\looseness=-1 Each category is formally defined and paired with a corresponding recovery action, ensuring that inefficiency detection translates into actionable supervisory guidance rather than generic critique. For example, in the case of \textit{task specification violation}, the prescribed recovery action is to redirect the agent to original task requirements. Full category definitions and recovery mappings are provided in Appendix~\ref{sec:prompts}.

\subsubsection{Guidance Generation}
At fixed intervals, the PRM is invoked to provide course-corrective feedback. Every $n$ steps, it receives as input: (i) the original problem description $d$, and (ii) the most recent $k$ steps of the agent’s transcript
\[
\mathcal{H}^{(k)}_t \;=\; \big(u_{t-k+1},a_{t-k+1},o_{t-k+1},\,\ldots,\,u_t,a_t,o_t\big),
\]
where $u_i$ are \emph{thoughts}, $a_i$ are \emph{actions}, and $o_i$ are the corresponding \emph{observations}. These elements are serialized into a structured text prompt:
\[
x_t = \mathrm{serialize}(d, \mathcal{H}^{(k)}_t).
\]

The PRM then produces natural language feedback
\[
g_t \;=\; f_\phi(x_t, \mathcal{T}),
\]
where $\mathcal{T}$ is the taxonomy of inefficiencies described in Section~\ref{sec:prm-course-corrector}. The taxonomy anchors the reasoning of the PRM: guidance is framed in terms of specific inefficiency categories (e.g., looping, redundant backtracking, subgoal drift), rather than unconstrained critique. Importantly, $g_t$ is expressed in natural language that the policy model can readily integrate into its own reasoning process.

\subsubsection{Variants}
\looseness=-1 We study different variants of \texttt{SWE-PRM} integration where the PRM provides natural language guidance to the policy model. Appendix~\ref{sec:prompts} lists the prompts corresponding to each variant. In the unified setting, where the PRM and the policy are instantiated by the same model, we vary three axes: (i) conciseness of feedback (Concise vs.\ Detailed), 
(ii) inclusion of an illustrative example (Example vs.\ No Example), and 
(iii) whether the PRM’s reasoning (taxonomy-based error analysis) is provided to the policy model alongside the overall guidance (Guidance+Reasoning vs.\ Guidance-only). 
This yields the set of conditions shown in Table~\ref{tab:variantmap}. We take \texttt{SWE-PRM$_{D}$} (taxonomy-guided, detailed, with example, guidance+reasoning) as the canonical variant, since it is the richest form of feedback and aligns most directly with the intended role of a PRM. Moreover, we also study a simple PRM variant that utilizes the model's inherent understanding of trajectory-level errors, i.e. \texttt{SWE-PRM$_{S}$}, along with explicitly stating the next action to be taken by the policy model as part of the PRM's guidance \texttt{SWE-PRM$_{DR}$}. 

\begin{table}[t]
\centering
\caption{\texttt{SWE-PRM} variants. 'Simple' involves using the model's inherent understanding of trajectory-level errors as opposed to seeding the reasoning with the taxonomy. 'Action Reco.' refers to explicitly stating the next action that the policy model should take.}
\label{tab:variantmap}
\begin{tabular}{lccccc}
\toprule
Name & Feedback Style & Example & Policy Input & Action Reco. \\
\midrule
\texttt{SWE-PRM$_{S}$}   & Simple & --  & Guidance+Reasoning & $\times$ \\
\texttt{SWE-PRM$_{C}$}   & Concise            & $\checkmark$ & Guidance+Reasoning & $\times$ \\
\texttt{SWE-PRM$_{CG}$}  & Concise            & $\checkmark$ & Guidance-only      & $\times$ \\
\texttt{SWE-PRM$_{D}$}   & Detailed           & $\checkmark$ & Guidance+Reasoning & $\times$ \\
\texttt{SWE-PRM$_{DN}$}  & Detailed           & $\times$     & Guidance+Reasoning & $\times$ \\
\texttt{SWE-PRM$_{DG}$}  & Detailed           & $\checkmark$ & Guidance-only      & $\times$ \\
\texttt{SWE-PRM$_{DNG}$} & Detailed           & $\times$     & Guidance-only      & $\times$ \\
\texttt{SWE-PRM$_{DR}$}  & Detailed           & $\checkmark$ & Guidance+Reasoning & $\checkmark$ \\
\bottomrule
\end{tabular}
\end{table}

In addition, we evaluate a subset of these settings with an expert PRM, where a stronger closed-source model provides guidance to a weaker open-source policy model. Specifically, we consider \texttt{SWE-PRM$_{S}$}, \texttt{SWE-PRM$_{D}$}, and \texttt{SWE-PRM$_{DR}$}, which capture the key baselines. We restrict the grid here due to the high cost of expert PRM queries, focusing on the most informative comparisons while keeping experiments tractable.

\section{Experimental Setup}

\subsection{Dataset}
We evaluate the proposed framework on \textsc{SWE-bench Verified}~\cite{jimenez2024swebench}, a subset of \textsc{SWE-bench} that has been verified by human annotators.  
As explained in Section~\ref{sec:task-arch}, the task involves repository-level bug fixing with long-horizon multi-step reasoning. The benchmark contains 500 instances paired with validated ground-truth patches. Unlike synthetic tasks, these instances reflect the complexity of real-world software engineering.  
The dataset serves as a standardized testbed for both baseline policies and PRM-supervised variants.

\subsection{Models and Hyperparameters} \label{sec:models}
\looseness=-1 We evaluate both open-source and proprietary models. Our experiments include three representative baselines for open-weights models: \textsc{SWE-agent-LM-32B} \footnote{\url{https://huggingface.co/SWE-bench/SWE-agent-LM-32B}}, \textsc{Devstral-Small-2505} \footnote{\url{https://huggingface.co/mistralai/Devstral-Small-2505}}, and \textsc{Devstral-Small-2507} \footnote{\url{https://huggingface.co/mistralai/Devstral-Small-2507}}, along with \textsc{Claude-Sonnet-4}.  
The \texttt{temperature} was set to $0.0$ for deterministic outputs for all models and the \texttt{top\_p} was set to $1.0$. For all experiments, we run the agent for a maximum of 75 steps, after which the run is auto-terminated and if a patch is generated, it is auto-submitted. For PRM-guided runs, we pass $k=8$ most recent steps and the PRM is invoked every $n=5$ steps. These hyperparameters balance contextual coverage with computational overhead and are fixed across all reported experiments. Two NVIDIA A100 GPUs were used to serve the models.

\subsection{Evaluation Metrics}
\paragraph{Resolution Rate.} The \% of instances correctly solved, both the overall rate and breakdowns by difficulty \cite{ganhotra2025difficulty}: (1) Easy ($\leq$15 minutes for human developers; 194 instances, 38.8\% of total), (2) Medium (15–60 minutes; 261 instances, 52.2\% of total), and (3) Hard ($\geq$1 hour; 45 instances, 9.0\% of total). This stratification highlights whether improvements generalize beyond the easiest cases.  

\paragraph{Patch Generation Rate.} The frequency with which a candidate patch is produced before the agent terminates, irrespective of correctness. This includes both, the patches submitted directly by the agent using the \texttt{submit} action, as well as auto-submissions in case of termination.
% This reflects an agent’s productivity and willingness to attempt solutions.  

\paragraph{Average Steps.} The average number of steps taken by the policy model per trajectory.  

% \paragraph{PRM engagement.}
% At fixed intervals (every $n$ steps), the PRM receives a sliding window of the most recent $k$ steps. For settings that seed PRM reasoning with the taxonomy (i.e. all except \texttt{PRM$_S$}), the PRM performs per-category detection (“Detected: Yes/No”). We count a window as \emph{suboptimal} if any error category is detected as “Yes”; otherwise it is \emph{optimal}. We also report average PRM invocations per trajectory (which scale with number of steps since $n$ is fixed).

\paragraph{Cost.} We report monetary cost in \$ per 100 instances, including the cost of running the policy model as well as the PRM interventions. For open source models, we consider API pricing from GPU cloud platforms \footnote{\url{https://www.together.ai/}} as of July 2025. (\$0.08 per million tokens).
For the closed source model, \textsc{Claude-Sonnet-4}, we consider API pricing as of July 2025 (\$ 3 and \$ 15 per million tokens for input and output respectively).  
% Token statistics are not shown separately, since they are already reflected in cost estimates.

\section{Results and Analysis}
\label{sec:results}

We evaluate the effectiveness of \texttt{SWE-PRM} across four dimensions: (i) their impact on overall resolution, (ii) performance stratified by task difficulty, (iii) the relative effectiveness of different feedback strategies, and (iv) the cost–benefit tradeoffs of using \texttt{SWE-PRM}. Unless otherwise noted, results are reported with \textsc{SWE-agent-LM-32B} as the base policy model. Full tables are provided in Appendix \ref{sec:full-results}; here we highlight the most salient results.

\begin{table}[t]
\caption{Open-Source \texttt{SWE-PRM} variations: \texttt{SWE-PRM} is same as policy model. $\Delta$s in brackets compare to the corresponding \texttt{base} row for each policy. Resolution rate $\Delta$s: \textcolor{ForestGreen}{green} = higher is better. Steps, Cost $\Delta$s: \textcolor{ForestGreen}{green} = lower is better. Numbers in \textbf{bold} are best for that model.}
\label{tab:open-prm}
\resizebox{\textwidth}{!}{%
\begin{tabular}{@{}llcccc@{}}
\toprule
\textbf{Setting} & \textbf{Policy Model} &
\multicolumn{1}{l}{\textbf{\begin{tabular}[c]{@{}l@{}}Resolution\\ Rate (\%)\end{tabular}}} &
\multicolumn{1}{l}{\textbf{\begin{tabular}[c]{@{}l@{}}Patch Generation\\ Rate (\%)\end{tabular}}} &
\multicolumn{1}{l}{\textbf{\begin{tabular}[c]{@{}l@{}}Avg Steps\end{tabular}}} &
\multicolumn{1}{l}{\textbf{\begin{tabular}[c]{@{}l@{}}Total Cost (\$) per\\100 instances\end{tabular}}} \\
\midrule
\multirow{3}{*}{\texttt{base}}
  & \textsc{SWE-agent-LM-32B}    & \textbf{40.0} & 92.4 & 38.64 & 2.77 \\
  & \textsc{Devstral-Small-2505} & 34.0 & 92.6 & 37.97 & \textbf{2.69} \\
  & \textsc{Devstral-Small-2507} & 30.0 & 88.0 & 40.16 & \textbf{2.70} \\
\midrule
\multirow{3}{*}{\texttt{SWE-PRM$_{S}$}}
  & \textsc{SWE-agent-LM-32B}    & 19.6 \textcolor{red}{(-20.4)} & 67.6 & 21.31 \textcolor{ForestGreen}{(-17.33)} & 2.46 \textcolor{ForestGreen}{(-0.31)} \\
  & \textsc{Devstral-Small-2505} & 34.4 \textcolor{ForestGreen}{(+0.4)} & 94.9 & 41.28 \textcolor{red}{(+3.31)} & 4.80 \textcolor{red}{(+2.11)} \\
  & \textsc{Devstral-Small-2507} & \textbf{33.6 \textcolor{ForestGreen}{(+3.6)}} & 93.4 & 45.54 \textcolor{red}{(+5.38)} & 4.84 \textcolor{red}{(+2.14)} \\
\midrule
\multirow{3}{*}{\texttt{SWE-PRM$_C$}}
  & \textsc{SWE-agent-LM-32B}    & 35.6 \textcolor{red}{(-4.4)} & 91.4 & 34.32 \textcolor{ForestGreen}{(-4.32)} & 3.77 \textcolor{red}{(+1.00)} \\
  & \textsc{Devstral-Small-2505} & 34.2 \textcolor{ForestGreen}{(+0.2)} & 92.2 & 38.39 \textcolor{red}{(+0.42)} & 3.96 \textcolor{red}{(+1.27)} \\
  & \textsc{Devstral-Small-2507} & 30.2 \textcolor{ForestGreen}{(+0.2)} & 90.2 & 43.46 \textcolor{red}{(+3.30)} & 4.46 \textcolor{red}{(+1.76)} \\
\midrule
\multirow{3}{*}{\texttt{SWE-PRM$_{CG}$}}
  & \textsc{SWE-agent-LM-32B}    & 35.6 \textcolor{red}{(-4.4)} & 89.8 & 32.71 \textcolor{ForestGreen}{(-5.93)} & 3.16 \textcolor{red}{(+0.39)} \\
  & \textsc{Devstral-Small-2505} & 34.2 \textcolor{ForestGreen}{(+0.2)} & 92.8 & 37.65 \textcolor{ForestGreen}{(-0.32)} & 3.27 \textcolor{red}{(+0.58)} \\
  & \textsc{Devstral-Small-2507} & 30.2 \textcolor{ForestGreen}{(+0.2)} & 91.0 & 41.52 \textcolor{red}{(+1.36)} & 3.73 \textcolor{red}{(+1.03)} \\
\midrule
\multirow{3}{*}{\texttt{SWE-PRM$_{D}$}}
  & \textsc{SWE-agent-LM-32B}    & 38.8 \textcolor{red}{(-1.2)} & 92.2 & 33.12 \textcolor{ForestGreen}{(-5.52)} & 3.31 \textcolor{red}{(+0.54)} \\
  & \textsc{Devstral-Small-2505} & 34.2 \textcolor{ForestGreen}{(+0.2)} & 93.4 & 37.89 \textcolor{ForestGreen}{(-0.08)} & 3.86 \textcolor{red}{(+1.17)} \\
  & \textsc{Devstral-Small-2507} & 30.2 \textcolor{ForestGreen}{(+0.2)} & 93.4 & 40.08 \textcolor{ForestGreen}{(-0.08)} & 4.15 \textcolor{red}{(+1.45)} \\
\midrule
\multirow{3}{*}{\texttt{SWE-PRM$_{DN}$}}
  & \textsc{SWE-agent-LM-32B}    & 30.0 \textcolor{red}{(-10.0)} & 79.6 & 27.54 \textcolor{ForestGreen}{(-11.10)} & 3.18 \textcolor{red}{(+0.41)} \\
  & \textsc{Devstral-Small-2505} & 34.2 \textcolor{ForestGreen}{(+0.2)} & 94.4 & 37.72 \textcolor{ForestGreen}{(-0.25)} & 4.06 \textcolor{red}{(+1.37)} \\
  & \textsc{Devstral-Small-2507} & 30.2 \textcolor{ForestGreen}{(+0.2)} & 91.6 & 39.98 \textcolor{ForestGreen}{(-0.18)} & 4.53 \textcolor{red}{(+1.83)} \\
\midrule
\multirow{3}{*}{\texttt{SWE-PRM$_{DG}$}}
  & \textsc{SWE-agent-LM-32B}    & 34.8 \textcolor{red}{(-5.2)} & \textbf{93.2} & 33.82 \textcolor{ForestGreen}{(-4.82)} & 2.97 \textcolor{red}{(+0.20)} \\
  & \textsc{Devstral-Small-2505} & 34.2 \textcolor{ForestGreen}{(+0.2)} & \textbf{95.4} & 38.58 \textcolor{red}{(+0.61)} & 3.47 \textcolor{red}{(+0.78)} \\
  & \textsc{Devstral-Small-2507} & 30.2 \textcolor{ForestGreen}{(+0.2)} & 93.0 & 39.52 \textcolor{ForestGreen}{(-0.64)} & 3.39 \textcolor{red}{(+0.69)} \\
\midrule
\multirow{3}{*}{\texttt{SWE-PRM$_{DNG}$}}
  & \textsc{SWE-agent-LM-32B}    & 30.0 \textcolor{red}{(-10.0)} & 54.8 & \textbf{10.11 \textcolor{ForestGreen}{(-28.53)}} & \textbf{1.23 \textcolor{ForestGreen}{(-1.54)}} \\
  & \textsc{Devstral-Small-2505} & 34.2 \textcolor{ForestGreen}{(+0.2)} & 94.4 & 36.05 \textcolor{ForestGreen}{(-1.92)} & 3.29 \textcolor{red}{(+0.60)} \\
  & \textsc{Devstral-Small-2507} & 30.4 \textcolor{ForestGreen}{(+0.4)} & 91.8 & 39.22 \textcolor{ForestGreen}{(-0.94)} & 3.38 \textcolor{red}{(+0.68)} \\
\midrule
\multirow{3}{*}{\texttt{SWE-PRM$_{DR}$}}
  & \textsc{SWE-agent-LM-32B}    & 36.8 \textcolor{red}{(-3.2)} & 92.8 & 28.67 \textcolor{ForestGreen}{(-9.97)} & 2.82 \textcolor{red}{(+0.05)} \\
  & \textsc{Devstral-Small-2505} & \textbf{36.0 \textcolor{ForestGreen}{(+2.0)}} & 95.0 & \textbf{32.33 \textcolor{ForestGreen}{(-5.64)}} & 3.06 \textcolor{red}{(+0.37)} \\
  & \textsc{Devstral-Small-2507} & 32.4 \textcolor{ForestGreen}{(+2.4)} & \textbf{94.4} & \textbf{37.67 \textcolor{ForestGreen}{(-2.49)}} & 3.87 \textcolor{red}{(+1.17)} \\
\bottomrule
\end{tabular}%
}
\end{table}

% % Please add the following required packages to your document preamble:
% % \usepackage{booktabs}
% % \usepackage{graphicx}
% \begin{table}[]
% \caption{Strong PRM variations}
% \label{tab:expert-variations}
% \resizebox{\columnwidth}{!}{%
% \begin{tabular}{@{}llrrrrrr@{}}
% \toprule
% \textbf{Setting} &
%   \textbf{Model} &
%   \multicolumn{1}{l}{\textbf{\begin{tabular}[c]{@{}l@{}}Resolution\\ Rate (\%)\end{tabular}}} &
%   \multicolumn{1}{l}{\textbf{\begin{tabular}[c]{@{}l@{}}Patch Generation\\ Rate (\%)\end{tabular}}} &
%   \multicolumn{1}{l}{\textbf{Avg Steps}} &
%   \multicolumn{1}{l}{\textbf{\begin{tabular}[c]{@{}l@{}}Policy Model\\ Cost (\$)\end{tabular}}} &
%   \multicolumn{1}{l}{\textbf{\begin{tabular}[c]{@{}l@{}}PRM\\ Cost (\$)\end{tabular}}} &
%   \multicolumn{1}{l}{\textbf{Total Cost (\$)}} \\ \midrule
% $base$                                      & \textsc{SWE-agent-LM-32B} & 40.0 & 92.4 & 38.64 & 0.0277 & 0      & 0.0277 \\
% $base$                                      & \textsc{Claude-Sonnet-4 }           & 66.6 & 100  & 61.72 & 1.2166 & 0      & 1.2166 \\
% $PRM_{\text{simple}}$                        & \textsc{SWE-agent-LM-32B} & 45.8 & 98.2 & 51.54 & 0.048  & 0.2362 & 0.2842 \\
% $PRM_{\text{detailed}}$             & \textsc{SWE-agent-LM-32B} & 50.6 & 98.2 & 37.99 & 0.0361 & 0.2237 & 0.2598 \\
% $PRM_{\text{detailed, rec. action}}$ & \textsc{SWE-agent-LM-32B} & 44.8 & 98.2 & 34.38 & 0.0316 & 0.2137 & 0.2453 \\ \bottomrule
% \end{tabular}%
% }
% \end{table}

% \usepackage{booktabs}
% \usepackage{graphicx}
\begin{table}[t]
\caption{Closed-Source \texttt{SWE-PRM} variations: \texttt{SWE-PRM} is \textsc{Claude-Sonnet-4} in all cases. Deltas in brackets compare to the \texttt{base} \textsc{SWE-agent-LM-32B} row.}
\label{tab:expert-variations}
\resizebox{\columnwidth}{!}{%
\begin{tabular}{@{}llcccc@{}}
\toprule
\textbf{Setting} &
  \textbf{Policy Model} &
  \multicolumn{1}{l}{\textbf{\begin{tabular}[c]{@{}l@{}}Resolution\\ Rate (\%)\end{tabular}}} &
  \multicolumn{1}{l}{\textbf{\begin{tabular}[c]{@{}l@{}}Patch Generation\\ Rate (\%)\end{tabular}}} &
  \multicolumn{1}{l}{\textbf{Avg Steps}} &
  \multicolumn{1}{l}{\textbf{\begin{tabular}[c]{@{}l@{}}Total Cost (\$) per\\100 instances\end{tabular}}} \\
\midrule
\multirow{2}{*}{\texttt{base}} 
   & \textsc{SWE-agent-LM-32B} & 40.0 & 92.4 & 38.64 & 2.77 \\
   & \textsc{Claude-Sonnet-4}  & 66.6 & 100.0 & 61.72 & 121.66 \\
\midrule
\texttt{SWE-PRM$_{S}$}  & \textsc{SWE-agent-LM-32B} & 45.8 \textcolor{ForestGreen}{(+5.8)} & \textbf{98.2} & 51.54 \textcolor{red}{(+12.90)} & 28.42 \textcolor{red}{(+25.65)}\\
\texttt{SWE-PRM$_{D}$}  & \textsc{SWE-agent-LM-32B} & \textbf{50.6 \textcolor{ForestGreen}{(+10.6)}} & \textbf{98.2} & 37.99 \textcolor{ForestGreen}{(-0.65)} & 25.98 \textcolor{red}{(+23.21)}\\
\texttt{SWE-PRM$_{DR}$} & \textsc{SWE-agent-LM-32B} & 44.8 \textcolor{ForestGreen}{(+4.8)}  & \textbf{98.2} & \textbf{34.38 \textcolor{ForestGreen}{(-4.26)}} &\textbf{ 24.53 \textcolor{red}{(+21.76)}} \\
\bottomrule
\end{tabular}%
}
\end{table}

\subsection{Do off-the-shelf \texttt{SWE-PRM}s improve performance over base agents?}
\label{sec:rq1}

\paragraph{Open-source \texttt{SWE-PRM} variants.}
Table~\ref{tab:open-prm} compares the base \textsc{SWE-agent-LM-32B} with six open-source PRM-guided configurations. None improve resolution consistently: the base achieves 40.0\% resolution, while open-source PRM variants range between 30.0--38.8\%. In addition, these variants often introduce inefficiencies such as longer trajectories or lower patch generation rates. Similarly, the \textsc{Devstral-small-2505} and \textsc{Devstral-small-2507} show little benefit from PRM guidance. These results suggest that models finetuned for SWE and agentic tasks are not inherently reliable when used as PRMs.

\paragraph{Closed-source PRM variants.}
In contrast, Table~\ref{tab:expert-variations} shows that PRMs based on \textsc{Claude-Sonnet-4} consistently raise resolution rates above the base. Improvements range from +4.8 to +10.6 percentage points, establishing a clear difference between open- and closed-source settings. The relative effectiveness of different feedback strategies is analyzed further in Section~\ref{sec:rq3}.

\paragraph{Takeaway.} Open-source PRMs fail to improve performance significantly over base agents, whereas closed-source PRMs consistently provide resolution gains of 5–11 percentage points.

\begin{figure*}[t]
    \centering
    \begin{subfigure}[t]{0.48\textwidth}
        \centering
        \includegraphics[width=\textwidth]{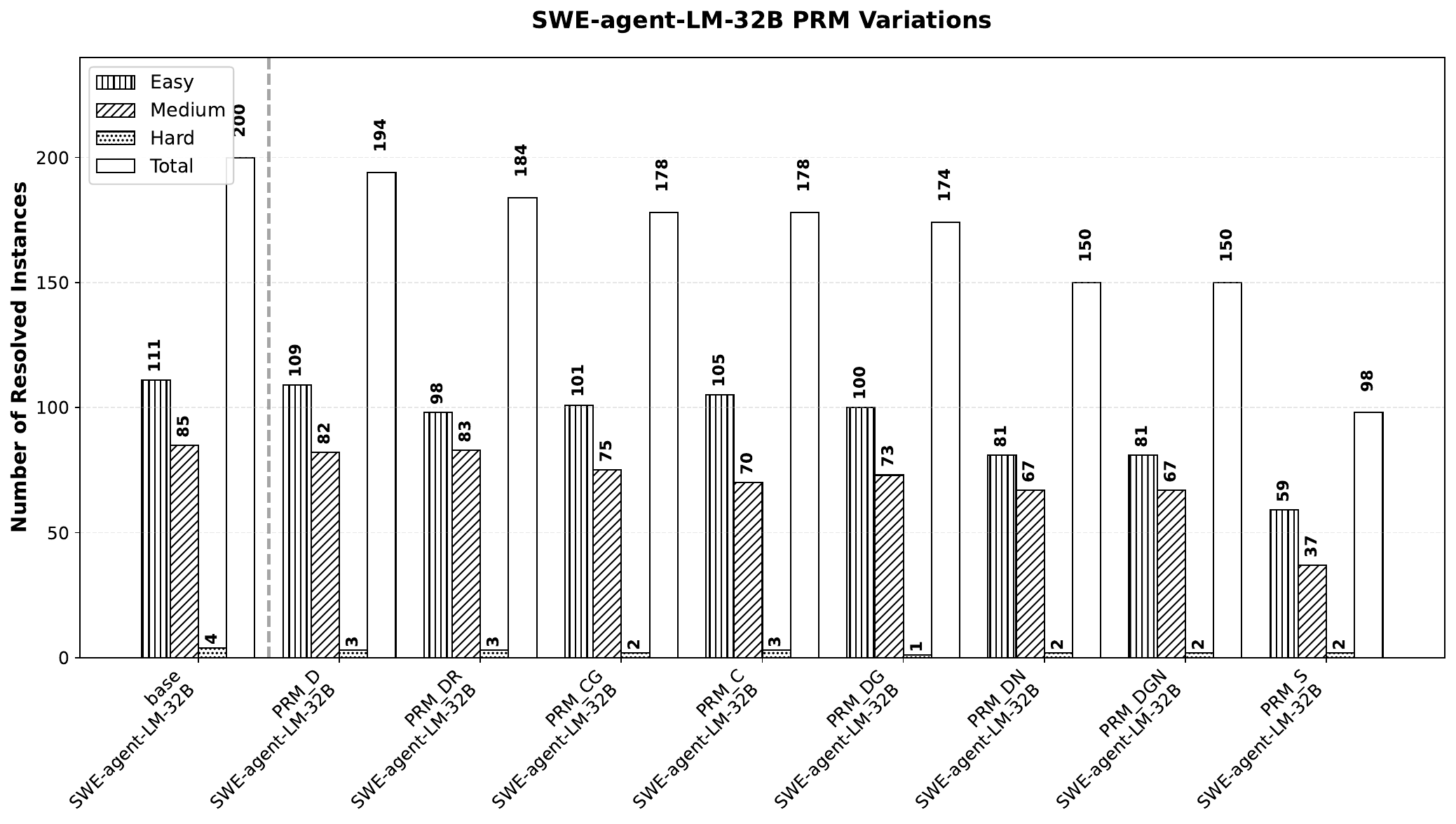}
        \caption{\textsc{SWE-agent-LM-32B} PRM variations}
        \label{fig:swe_agent_variations}
    \end{subfigure}
    \hfill
    \begin{subfigure}[t]{0.48\textwidth}
        \centering
        \includegraphics[width=\textwidth]{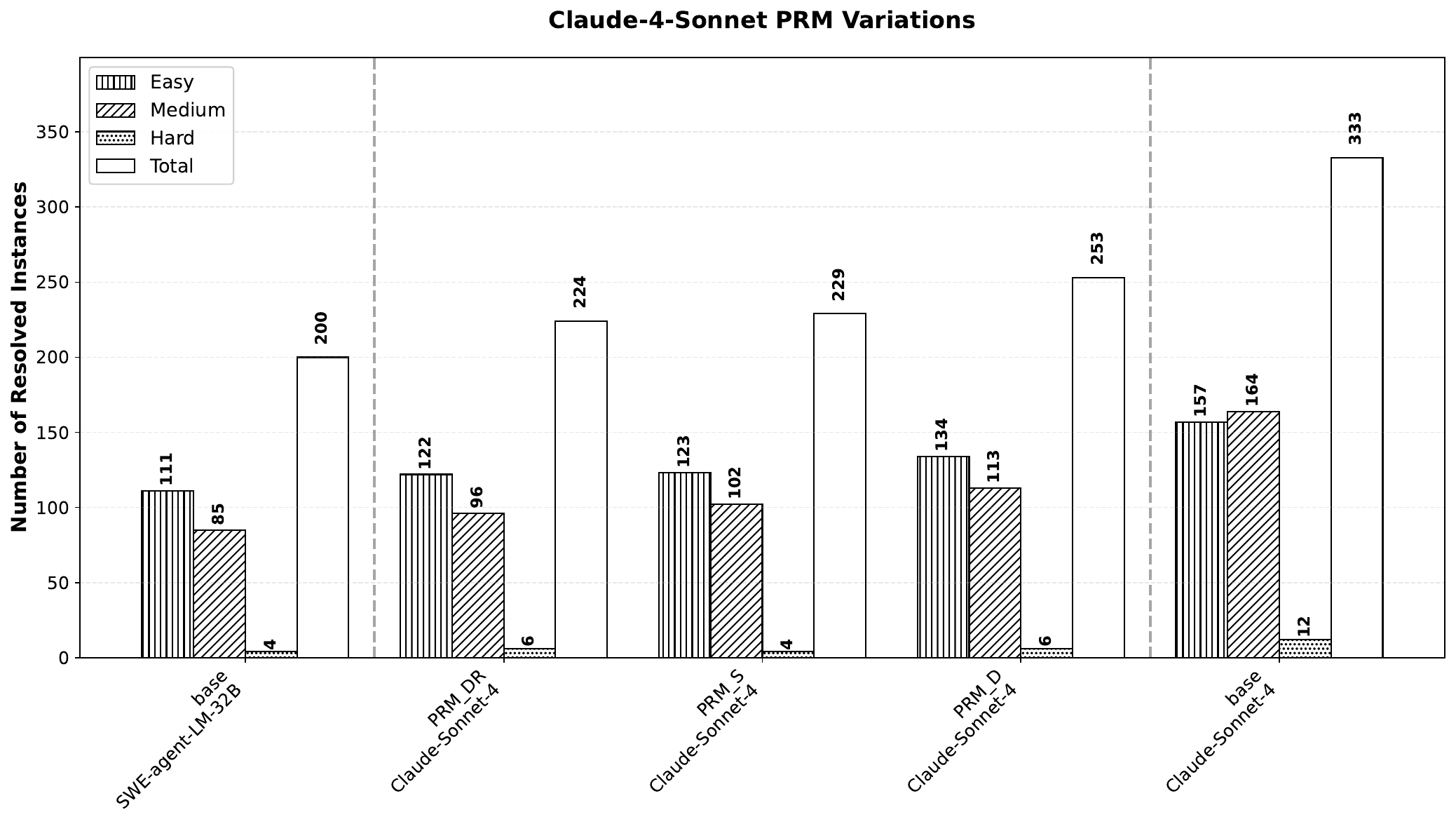}
        \caption{\textsc{Claude-Sonnet-4} PRM variations}
        \label{fig:claude_variations}
    \end{subfigure}
    \caption{Difficulty-wise instances resolved out of 500 SWE-bench Verified instances (194 Easy, 261 Medium, 45 Hard). \texttt{PRM$_D$} with \textsc{Claude-Sonnet-4} yields the strongest gains across all tiers.}
    \label{fig:prm_comparison}
\end{figure*}

\subsection{How does performance vary across difficulty levels?}
\label{sec:rq2}

We focus on \textsc{SWE-agent-LM-32B} for difficulty-stratified analysis, as it achieves the best base performance among the open-source models (40.0\% resolution overall). Figure~\ref{fig:prm_comparison} shows results across Easy (194), Medium (261), and Hard (45) instances. The base agent achieves 57.2\% on Easy, 32.6\% on Medium, and only 8.9\% on Hard, indicating a steep performance drop on more complex tasks.
Open-source PRM variants (Figure~\ref{fig:swe_agent_variations}) do not improve this distribution. For example, \texttt{PRM$_C$} and \texttt{PRM$_{CG}$} reduce overall resolution, while \texttt{PRM$_{DN}$} and \texttt{PRM$_{DGN}$} degrade Hard-task performance further.  
Closed-source PRMs with \textsc{Claude-Sonnet-4} (Figure~\ref{fig:claude_variations}) improve across all tiers. The strongest setting, \texttt{PRM$_D$}, reaches 69.1\% on Easy, 43.3\% on Medium, and 13.3\% on Hard. Even unguided reasoning (\texttt{PRM$_S$}) improves every tier, though it lengthens trajectories. These gains show that PRMs are particularly valuable for Medium and Hard tasks, where trajectory-level inefficiencies are most damaging.

\paragraph{Takeaway.} Open-source PRMs provide no benefit across difficulty levels, while closed-source PRMs, especially \texttt{PRM$_D$}, deliver consistent improvements, with the largest relative gains on Medium and Hard tasks.

\subsection{Which course correction strategies are most effective?}
\label{sec:rq3}

We next individually compare three feedback strategies with \textsc{Claude-Sonnet-4}: simple unguided reasoning (\texttt{PRM$_S$}), detailed taxonomy-guided reasoning with feedback (\texttt{PRM$_D$}), and detailed taxonomy-guided reasoning with explicit action recommendation (\texttt{PRM$_{DR}$}).  

\textbf{Unguided reasoning (\texttt{PRM$_S$})} improves resolution to 45.8\% ($+5.8$\,pp) but lengthens trajectories substantially (51.5 steps vs.\ 38.6 for base). Since no error detection is elicited, windows may not be explicitly flagged as suboptimal, providing no concrete signal about inefficient behavior; the empirical effect is longer, less efficient runs.  

\looseness=-1 \textbf{Taxonomy-guided feedback (\texttt{PRM$_D$})} is the strongest setting: resolution reaches 50.6\% ($+10.6$\,pp) while steps slightly decrease (37.99). Appendix Table \ref{tab:full-results-table} shows that nearly every PRM invocation marks the window as suboptimal (7.21 out of 7.24), indicating frequent detection of trajectory-level errors. This shows that structured signals help the agent truncate inefficient exploration rather than extend it.  

\textbf{Taxonomy-guided with action recommendation (\texttt{PRM$_{DR}$})} achieves the smallest resolution gain (44.8\%, $+4.8$\,pp). While steps reduce to 34.4, almost every invocation is still flagged suboptimal (6.37 out of 6.39), suggesting that rigid prescriptions lead to shorter but less successful runs.  

Across settings, closed-source PRM variants almost always flag windows as suboptimal, reflecting strong detection of trajectory-level issues. Open-source PRMs also mark windows as suboptimal, but at lower rates, aligning with their weaker overall effectiveness.
Taken together, these results demonstrate that taxonomy grounding is essential for effective guidance, and that providing explicit actions can harm resolution by constraining the agent too tightly.

\paragraph{Takeaway.} \texttt{PRM$_D$} is the most effective strategy, delivering the largest resolution rate gain with fewer steps; \texttt{PRM$S$} lengthens runs for limited benefit, and \texttt{PRM${DR}$} shortens runs but reduces accuracy.

\subsection{What are the cost–benefit tradeoffs of PRMs?}
\label{sec:rq4}

\looseness=-1 The final question is whether the substantial performance gains enabled by PRMs justify their additional inference cost. Table~\ref{tab:expert-variations} reports cost per 100 instances. The base \textsc{SWE-agent-LM-32B} resolves 40.0\% of instances at a cost of \$2.77. In contrast, closed-source PRMs increase resolution to as high as 50.6\%, a double-digit relative improvement, while raising cost to \$24--\$28 per 100 instances.

\looseness=-1 Breaking costs down by component in Appendix \ref{sec:full-results} shows that the increase is driven primarily by PRM queries: for example, \texttt{PRM$_D$} spends \$3.61 per 100 on policy calls and \$22.4 on PRM calls. Crucially, this overhead translates into more instances successfully resolved. Measured as incremental cost per additional success, \texttt{PRM$_D$} achieves the best tradeoff: \$23.2 in added cost yields 10.6 additional resolutions. \texttt{PRM$_S$} and \texttt{PRM$_{DR}$} are less favorable, but still surpass the base agent in absolute performance.

\looseness=-1 Viewed from this perspective, PRMs represent a deliberate performance–cost tradeoff. Without them, resolution plateaus at 40\%. With taxonomy-guided feedback (\texttt{PRM$_D$}), resolution climbs above 50\%. These results underscore that PRMs are a viable and practical means of unlocking further progress on complex tasks like repository-level code generation, and point to future work on making PRM calls more cost-efficient.

\looseness=-1 \paragraph{Takeaway.} PRMs are not a free improvement, but they deliver clear performance gains: \texttt{PRM$_D$} surpasses 50\% resolution and offers the best cost-benefit profile, making it the most effective path to higher accuracy today.

\section{Discussion and Conclusion}

This work introduces \texttt{SWE-PRM}, a real-time course-corrector for software engineering agents. By anchoring feedback in a taxonomy of trajectory-level inefficiencies, \texttt{SWE-PRM} delivers lightweight interventions that improves agent reliability without altering the base policy model. Our results on \textsc{SWE-bench Verified} demonstrate three key findings. First, while open-source PRMs offer little benefit, closed-source PRMs consistently boost resolution by 5-11 percentage points. Second, the strongest gains occur on medium and hard tasks, where trajectory-level inefficiencies are most pronounced. Third, among feedback strategies, taxonomy-guided PRMs provide the best balance: they improve the resolution rate to above 50\% while maintaining or reducing the trajectory lengths.

Beyond these results, our study highlights broader implications. PRMs shift the design space from purely outcome-focused optimization toward process-aware guidance, complementing approaches like search-based planning or post-hoc trajectory analysis. Although PRMs add inference overhead, their modularity allows them to be flexibly integrated with both open-weight and proprietary models. Future work could reduce costs through adaptive invocation schedules or distillation into lighter models and extend the taxonomy to other sequential reasoning domains. 
In sum, PRMs represent a practical and principled path forward: they enable agents to not only solve more tasks, but to solve them more efficiently, setting the stage for more reliable deployment of LLM agents in complex software engineering environments.

\bibliographystyle{plainnat} 
\bibliography{refs} 

\appendix
\clearpage
\section{Appendix}

\subsection{Prompts}\label{sec:prompts}

\begin{lstlisting}[
  style=promptstyle,
  caption={Common instructions used for all runs},
  label={lst:instructions-prompt}
]
system_template: |-
      You are a helpful assistant that can interact with a computer to solve tasks.
      <IMPORTANT>
      * If user provides a path, you should NOT assume it's relative to the current working directory. Instead, you should explore the file system to find the file before working on it.
      </IMPORTANT>

      You have access to the following functions:

      ---- BEGIN FUNCTION #1: bash ----
      Description: Execute a bash command in the terminal.

      Parameters:
        (1) command (string, required): The bash command to execute. Can be empty to view additional logs when previous exit code is `-1`. Can be `ctrl+c` to interrupt the currently running process.
      ---- END FUNCTION #1 ----

      ---- BEGIN FUNCTION #2: submit ----
      Description: Finish the interaction when the task is complete OR if the assistant cannot proceed further with the task.
      No parameters are required for this function.
      ---- END FUNCTION #2 ----

      ---- BEGIN FUNCTION #3: str_replace_editor ----
      Description: Custom editing tool for viewing, creating and editing files
      * State is persistent across command calls and discussions with the user
      * If `path` is a file, `view` displays the result of applying `cat -n`. If `path` is a directory, `view` lists non-hidden files and directories up to 2 levels deep
      * The `create` command cannot be used if the specified `path` already exists as a file
      * If a `command` generates a long output, it will be truncated and marked with `<response clipped>`
      * The `undo_edit` command will revert the last edit made to the file at `path`

      Notes for using the `str_replace` command:
      * The `old_str` parameter should match EXACTLY one or more consecutive lines from the original file. Be mindful of whitespaces!
      * If the `old_str` parameter is not unique in the file, the replacement will not be performed. Make sure to include enough context in `old_str` to make it unique
      * The `new_str` parameter should contain the edited lines that should replace the `old_str`

      Parameters:
        (1) command (string, required): The commands to run. Allowed options are: `view`, `create`, `str_replace`, `insert`, `undo_edit`.
      Allowed values: [`view`, `create`, `str_replace`, `insert`, `undo_edit`]
        (2) path (string, required): Absolute path to file or directory, e.g. `/repo/file.py` or `/repo`.
        (3) file_text (string, optional): Required parameter of `create` command, with the content of the file to be created.
        (4) old_str (string, optional): Required parameter of `str_replace` command containing the string in `path` to replace.
        (5) new_str (string, optional): Optional parameter of `str_replace` command containing the new string (if not given, no string will be added). Required parameter of `insert` command containing the string to insert.
        (6) insert_line (integer, optional): Required parameter of `insert` command. The `new_str` will be inserted AFTER the line `insert_line` of `path`.
        (7) view_range (array, optional): Optional parameter of `view` command when `path` points to a file. If none is given, the full file is shown. If provided, the file will be shown in the indicated line number range, e.g. [11, 12] will show lines 11 and 12. Indexing at 1 to start. Setting `[start_line, -1]` shows all lines from `start_line` to the end of the file.
      ---- END FUNCTION #3 ----


      If you choose to call a function ONLY reply in the following format with NO suffix:

      Provide any reasoning for the function call here.
      <function=example_function_name>
      <parameter=example_parameter_1>value_1</parameter>
      <parameter=example_parameter_2>
      This is the value for the second parameter
      that can span
      multiple lines
      </parameter>
      </function>

      <IMPORTANT>
      Reminder:
      - Function calls MUST follow the specified format, start with <function= and end with </function>
      - Required parameters MUST be specified
      - Only call one function at a time
      - Always provide reasoning for your function call in natural language BEFORE the function call (not after)
      </IMPORTANT>
instance_template: |-
      <uploaded_files>
      {{working_dir}}
      </uploaded_files>
      I've uploaded a python code repository in the directory {{working_dir}}. Consider the following PR description:

      <pr_description>
      {{problem_statement}}
      </pr_description>

      Can you help me implement the necessary changes to the repository so that the requirements specified in the <pr_description> are met?
      I've already taken care of all changes to any of the test files described in the <pr_description>. This means you DON'T have to modify the testing logic or any of the tests in any way!
      Your task is to make the minimal changes to non-tests files in the {{working_dir}} directory to ensure the <pr_description> is satisfied.
      Follow these steps to resolve the issue:
      1. As a first step, it might be a good idea to find and read code relevant to the <pr_description>
      2. Create a script to reproduce the error and execute it with `python <filename.py>` using the bash tool, to confirm the error
      3. Edit the source code of the repo to resolve the issue
      4. Rerun your reproduce script and confirm that the error is fixed!
      5. Think about edgecases and make sure your fix handles them as well
      Your thinking should be thorough and so it's fine if it's very long.
next_step_template: |-
      OBSERVATION:
      {{observation}}
next_step_no_output_template: |-
      Your command ran successfully and did not produce any output.
\end{lstlisting}

\begin{lstlisting}[
  style=promptstyle,
  caption={Prompt for the \texttt{PRM$_S$} variant.},
  label={lst:prm_s-prompt}
]
You are an expert at analyzing SWE-agent trajectories for solving software engineering tasks. Your role is to detect trajectory-level errors and provide corrective guidance to prevent task failure.
Analyze these recent agent steps for suboptimality. 

Common suboptimal patterns include:
1. **Repetitive actions**: Viewing the same file multiple times without making progress, running the same commands repeatedly
2. **Not understanding observations**: Ignoring error messages, missing important information in outputs, not reading test failures carefully
3. **Inefficient exploration**: Looking at irrelevant files, not focusing on the error location, exploring too broadly
4. **Missing obvious next steps**: Not running tests after changes, not checking if changes work, not reading error traces
5. **Redundant operations**: Creating files that exist, re-implementing existing functionality
6. **Poor debugging strategy**: Making random changes without understanding the issue, not isolating the problem
7. **Incomplete actions**: Starting to edit but not finishing, viewing files without purpose

Recent agent steps:

{{recent_steps}}

If the recent steps are suboptimal, provide DETAILED, ACTIONABLE course-corrective feedback that:
- Explains specifically why these steps are problematic given the context
- Suggests concrete next steps the agent should take instead
- References specific files, functions, or error messages from the trajectory
- Provides a clear strategy to get back on track
- Is at least 3-5 sentences long with specific technical guidance
\end{lstlisting}

\begin{lstlisting}[
  style=promptstyle,
  caption={Prompt for the \texttt{PRM$_C$} and \texttt{PRM$_{CG}$} variants.},
  label={lst:prm_c-prompt}
]
You are a supervisor monitoring an LLM-based coding agent. Your role is to detect trajectory-level errors and provide corrective guidance to prevent task failure.

Analyze the provided agent trace for the following error categories. For each detected error, provide specific evidence and a recovery action.

SPECIFICATION ERRORS (System Design Issues):
- Task Specification Violations: Agent fails to adhere to task constraints or requirements  
- Role Specification Violations: Agent behaves outside its defined role/responsibilities
- Step Repetition: Unnecessary repetition of completed steps or actions
- Termination Condition Unawareness: Agent continues working when task completion criteria are met

REASONING ERRORS (Decision Making Issues):
- Problem Misidentification: Agent misunderstands the core problem or current subtask
- Tool Selection Errors: Agent uses inappropriate tools for the current task
- Hallucinations: Agent generates false information or fabricates tool outputs
- Information Processing Failures: Poor retrieval of relevant information or misinterpretation

COORDINATION ERRORS (Process Management Issues):
- Task Derailment: Agent deviates from intended objective or loses focus
- Goal Deviation: Agent pursues goals that don't serve the main objective
- Context Handling Failures: Agent loses important context or forgets previous findings
- Verification Failures: Inadequate checking of work quality or correctness

For each error category, respond with:
DETECTED: Yes/No
EVIDENCE: Specific quote or observation from trace (if detected)
RECOVERY_ACTION: Specific instruction to correct the error (if detected)

Then provide:
TASK_STATUS: On track / Needs correction / Critical intervention required
OVERALL_GUIDANCE: 1-2 sentences of specific guidance for the agent

Recent agent steps:
{{recent_steps}}

Focus on errors that can be corrected through guidance. Be concise but precise in evidence citations. Only mark "DETECTED: Yes" if you have clear evidence.
\end{lstlisting}

\begin{lstlisting}[
  style=promptstyle,
  caption={Prompt for the \texttt{PRM$_D$} and \texttt{PRM$_{DG}$} variants.},
  label={lst:prm_d-prompt}
]
You are a supervisor monitoring an LLM-based coding agent. Your role is to detect trajectory-level errors and provide corrective guidance to prevent task failure.
Analyze the provided agent trace for the following error categories. For each detected error, provide specific evidence and a recovery action.

SPECIFICATION ERRORS (System Design Issues)

1. Task Specification Violations
Definition: Agent fails to adhere to task constraints or requirements
Recovery: Redirect agent to original task requirements

2. Role Specification Violations
Definition: Agent behaves outside its defined role/responsibilities
Recovery: Remind agent of its specific role and boundaries

3. Step Repetition
Definition: Unnecessary repetition of completed steps or actions
Recovery: Acknowledge completed work and guide to next logical step

4. Termination Condition Unawareness
Definition: Agent continues working when task completion criteria are met
Recovery: Signal completion criteria and instruct to finalize

REASONING ERRORS (Decision Making Issues)

5. Problem Misidentification
Definition: Agent misunderstands the core problem or current subtask
Recovery: Clarify the actual problem and expected approach

6. Tool Selection Errors
Definition: Agent uses inappropriate tools for the current task
Recovery: Suggest correct tools and explain their appropriate usage

7. Hallucinations
Definition: Agent generates false information or fabricates tool outputs
Recovery: Request verification of claims against actual evidence

8. Information Processing Failures
Definition: Poor retrieval of relevant information or misinterpretation
Recovery: Guide agent to correct information sources and interpretation

COORDINATION ERRORS (Process Management Issues)

9. Task Derailment
Definition: Agent deviates from intended objective or loses focus
Recovery: Realign agent with original objectives and priorities

10. Goal Deviation
Definition: Agent pursues goals that don't serve the main objective
Recovery: Refocus on primary goals and expected outcomes

11. Context Handling Failures
Definition: Agent loses important context or forgets previous findings
Recovery: Provide context summary and key information recap

12. Verification Failures
Definition: Inadequate checking of work quality or correctness
Recovery: Instruct specific verification steps and quality checks

Response Format

For each error category, respond with:
DETECTED: Yes/No
EVIDENCE: Specific quote or observation from trace (if detected)
RECOVERY_ACTION: Specific instruction to correct the error (if detected)

Then provide:
TASK_STATUS: On track / Needs correction / Critical intervention required
OVERALL_GUIDANCE: Detailed and specific guidance for the agent

Example Response Structure

SPECIFICATION ERRORS:
1. Task Specification Violations: DETECTED: No
2. Role Specification Violations: DETECTED: No  
3. Step Repetition: DETECTED: Yes
EVIDENCE: "Agent ran the same test command three times: 'pytest test_file.py'"
RECOVERY_ACTION: "The test has already been executed successfully. Proceed to analyze the results and move to the next development step."
4. Termination Condition Unawareness: DETECTED: No

REASONING ERRORS:
5. Problem Misidentification: DETECTED: No
6. Tool Selection Errors: DETECTED: Yes
EVIDENCE: "Agent used text editor to run Python code instead of using the Python interpreter"
RECOVERY_ACTION: "Use the Python interpreter tool for code execution. The text editor is for viewing and modifying files only."
7. Hallucinations: DETECTED: No
8. Information Processing Failures: DETECTED: No

COORDINATION ERRORS:
9. Task Derailment: DETECTED: No
10. Goal Deviation: DETECTED: No
11. Context Handling Failures: DETECTED: No
12. Verification Failures: DETECTED: No

TASK_STATUS: Needs correction
OVERALL_GUIDANCE: You are repeating actions unnecessarily and using incorrect tools. Specifically:
1. Stop running the same test command repeatedly - the test 'pytest test_file.py' has already been executed successfully three times with the same result
2. Use the Python interpreter tool for executing Python code, not the text editor which is only for viewing and modifying files
3. Now focus on analyzing the test results you already obtained to determine what the next development step should be
4. Review the test output to identify any failing tests or areas that need improvement
5. If all tests are passing, proceed to verify your implementation meets the original requirements before considering the task complete

Recent agent steps:

{{recent_steps}}

Instructions:

1. Focus on errors that can be corrected through guidance
2. Provide specific, actionable recovery instructions
3. Be concise but precise in evidence citations
4. Only mark "DETECTED: Yes" if you have clear evidence
5. Prioritize errors that most threaten task completion
\end{lstlisting}

\begin{lstlisting}[
  style=promptstyle,
  caption={Prompt for the \texttt{PRM$_{DN}$} and \texttt{PRM$_{DNG}$} variants.},
  label={lst:prm_dn-prompt}
]
You are a supervisor monitoring an LLM-based coding agent. Your role is to detect trajectory-level errors and provide corrective guidance to prevent task failure.
Analyze the provided agent trace for the following error categories. For each detected error, provide specific evidence and a recovery action.

SPECIFICATION ERRORS (System Design Issues)

1. Task Specification Violations
Definition: Agent fails to adhere to task constraints or requirements
Recovery: Redirect agent to original task requirements

2. Role Specification Violations
Definition: Agent behaves outside its defined role/responsibilities
Recovery: Remind agent of its specific role and boundaries

3. Step Repetition
Definition: Unnecessary repetition of completed steps or actions
Recovery: Acknowledge completed work and guide to next logical step

4. Termination Condition Unawareness
Definition: Agent continues working when task completion criteria are met
Recovery: Signal completion criteria and instruct to finalize

REASONING ERRORS (Decision Making Issues)

5. Problem Misidentification
Definition: Agent misunderstands the core problem or current subtask
Recovery: Clarify the actual problem and expected approach

6. Tool Selection Errors
Definition: Agent uses inappropriate tools for the current task
Recovery: Suggest correct tools and explain their appropriate usage

7. Hallucinations
Definition: Agent generates false information or fabricates tool outputs
Recovery: Request verification of claims against actual evidence

8. Information Processing Failures
Definition: Poor retrieval of relevant information or misinterpretation
Recovery: Guide agent to correct information sources and interpretation

COORDINATION ERRORS (Process Management Issues)

9. Task Derailment
Definition: Agent deviates from intended objective or loses focus
Recovery: Realign agent with original objectives and priorities

10. Goal Deviation
Definition: Agent pursues goals that don't serve the main objective
Recovery: Refocus on primary goals and expected outcomes

11. Context Handling Failures
Definition: Agent loses important context or forgets previous findings
Recovery: Provide context summary and key information recap

12. Verification Failures
Definition: Inadequate checking of work quality or correctness
Recovery: Instruct specific verification steps and quality checks

Response Format

For each error category, respond with:
DETECTED: Yes/No
EVIDENCE: Specific quote or observation from trace (if detected)
RECOVERY_ACTION: Specific instruction to correct the error (if detected)

Then provide:
TASK_STATUS: On track / Needs correction / Critical intervention required
OVERALL_GUIDANCE: Detailed and specific guidance for the agent

Recent agent steps:

{{recent_steps}}

Instructions:

1. Focus on errors that can be corrected through guidance
2. Provide specific, actionable recovery instructions
3. Be concise but precise in evidence citations
4. Only mark "DETECTED: Yes" if you have clear evidence
5. Prioritize errors that most threaten task completion
\end{lstlisting}

\begin{lstlisting}[
  style=promptstyle,
  caption={Prompt for the \texttt{PRM$_{DR}$} variant.},
  label={lst:prm_dr-prompt}
]
You are a supervisor monitoring an LLM-based coding agent. Your role is to detect trajectory-level errors and provide corrective guidance to prevent task failure. 

The agent has access to the following functions as actions - 

---- BEGIN FUNCTION #1: bash ----
Description: Execute a bash command in the terminal.

Parameters:
(1) command (string, required): The bash command to execute. Can be empty to view additional logs when previous exit code is `-1`. Can be `ctrl+c` to interrupt the currently running process.
---- END FUNCTION #1 ----

---- BEGIN FUNCTION #2: submit ----
Description: Finish the interaction when the task is complete OR if the assistant cannot proceed further with the task.
No parameters are required for this function.
---- END FUNCTION #2 ----

---- BEGIN FUNCTION #3: str_replace_editor ----
Description: Custom editing tool for viewing, creating and editing files
* State is persistent across command calls and discussions with the user
* If `path` is a file, `view` displays the result of applying `cat -n`. If `path` is a directory, `view` lists non-hidden files and directories up to 2 levels deep
* The `create` command cannot be used if the specified `path` already exists as a file
* If a `command` generates a long output, it will be truncated and marked with `<response clipped>`
* The `undo_edit` command will revert the last edit made to the file at `path`

Notes for using the `str_replace` command:
* The `old_str` parameter should match EXACTLY one or more consecutive lines from the original file. Be mindful of whitespaces!
* If the `old_str` parameter is not unique in the file, the replacement will not be performed. Make sure to include enough context in `old_str` to make it unique
* The `new_str` parameter should contain the edited lines that should replace the `old_str`

Parameters:
(1) command (string, required): The commands to run. Allowed options are: `view`, `create`, `str_replace`, `insert`, `undo_edit`.
Allowed values: [`view`, `create`, `str_replace`, `insert`, `undo_edit`]
(2) path (string, required): Absolute path to file or directory, e.g. `/repo/file.py` or `/repo`.
(3) file_text (string, optional): Required parameter of `create` command, with the content of the file to be created.
(4) old_str (string, optional): Required parameter of `str_replace` command containing the string in `path` to replace.
(5) new_str (string, optional): Optional parameter of `str_replace` command containing the new string (if not given, no string will be added). Required parameter of `insert` command containing the string to insert.
(6) insert_line (integer, optional): Required parameter of `insert` command. The `new_str` will be inserted AFTER the line `insert_line` of `path`.
(7) view_range (array, optional): Optional parameter of `view` command when `path` points to a file. If none is given, the full file is shown. If provided, the file will be shown in the indicated line number range, e.g. [11, 12] will show lines 11 and 12. Indexing at 1 to start. Setting `[start_line, -1]` shows all lines from `start_line` to the end of the file.
---- END FUNCTION #3 ----

Analyze the provided agent trace for the following error categories. For each detected error, provide specific evidence and a recovery action.

SPECIFICATION ERRORS (System Design Issues)

1. Task Specification Violations
Definition: Agent fails to adhere to task constraints or requirements
Recovery: Redirect agent to original task requirements

2. Role Specification Violations
Definition: Agent behaves outside its defined role/responsibilities
Recovery: Remind agent of its specific role and boundaries

3. Step Repetition
Definition: Unnecessary repetition of completed steps or actions
Recovery: Acknowledge completed work and guide to next logical step

4. Termination Condition Unawareness
Definition: Agent continues working when task completion criteria are met
Recovery: Signal completion criteria and instruct to finalize

REASONING ERRORS (Decision Making Issues)

5. Problem Misidentification
Definition: Agent misunderstands the core problem or current subtask
Recovery: Clarify the actual problem and expected approach

6. Tool Selection Errors
Definition: Agent uses inappropriate tools for the current task
Recovery: Suggest correct tools and explain their appropriate usage

7. Hallucinations
Definition: Agent generates false information or fabricates tool outputs
Recovery: Request verification of claims against actual evidence

8. Information Processing Failures
Definition: Poor retrieval of relevant information or misinterpretation
Recovery: Guide agent to correct information sources and interpretation

COORDINATION ERRORS (Process Management Issues)

9. Task Derailment
Definition: Agent deviates from intended objective or loses focus
Recovery: Realign agent with original objectives and priorities

10. Goal Deviation
Definition: Agent pursues goals that don't serve the main objective
Recovery: Refocus on primary goals and expected outcomes

11. Context Handling Failures
Definition: Agent loses important context or forgets previous findings
Recovery: Provide context summary and key information recap

12. Verification Failures
Definition: Inadequate checking of work quality or correctness
Recovery: Instruct specific verification steps and quality checks

Response Format

For each error category, respond with:
DETECTED: Yes/No
EVIDENCE: Specific quote or observation from trace (if detected)
RECOVERY_ACTION: Specific instruction to correct the error (if detected)

Then provide:
TASK_STATUS: On track / Needs correction / Critical intervention required
OVERALL_GUIDANCE: Detailed and specific guidance for the agent
RECOMMENDED_ACTION: Recommended next action that the agent should take

Example Response Structure

SPECIFICATION ERRORS:
1. Task Specification Violations: DETECTED: No
2. Role Specification Violations: DETECTED: No  
3. Step Repetition: DETECTED: Yes
EVIDENCE: "Agent ran the same test command three times: 'pytest test_file.py'"
RECOVERY_ACTION: "The test has already been executed successfully. Proceed to analyze the results and move to the next development step."
4. Termination Condition Unawareness: DETECTED: No

REASONING ERRORS:
5. Problem Misidentification: DETECTED: No
6. Tool Selection Errors: DETECTED: Yes
EVIDENCE: "Agent used text editor to run Python code instead of using the Python interpreter"
RECOVERY_ACTION: "Use the Python interpreter tool for code execution. The text editor is for viewing and modifying files only."
7. Hallucinations: DETECTED: No
8. Information Processing Failures: DETECTED: No

COORDINATION ERRORS:
9. Task Derailment: DETECTED: No
10. Goal Deviation: DETECTED: No
11. Context Handling Failures: DETECTED: No
12. Verification Failures: DETECTED: No

TASK_STATUS: Needs correction
OVERALL_GUIDANCE: You are repeating actions unnecessarily and using incorrect tools. Specifically:
1. Stop running the same test command repeatedly - the test 'pytest test_file.py' has already been executed successfully three times with the same result
2. Use the Python interpreter tool for executing Python code, not the text editor which is only for viewing and modifying files
3. Now focus on analyzing the test results you already obtained to determine what the next development step should be
4. Review the test output to identify any failing tests or areas that need improvement
5. If all tests are passing, proceed to verify your implementation meets the original requirements before considering the task complete
RECOMMENDED_ACTION: str_replace_editor view /path/to/test_output.log

Recent agent steps:

{{recent_steps}}

Instructions:

1. Focus on errors that can be corrected through guidance
2. Provide specific, actionable recovery instructions
3. Be concise but precise in evidence citations
4. Only mark "DETECTED: Yes" if you have clear evidence
5. Prioritize errors that most threaten task completion
6. Provide a concrete recommended next action for the agent to take. This should be from the functions available to the agent.
\end{lstlisting}

\subsection{Complete Results}\label{sec:full-results}

{\tiny
% \usepackage{booktabs}
% \usepackage{graphicx}
% \usepackage{lscape}
\begin{landscape}
\begin{table}[]
\centering
\caption{All metrics for all SWE-PRM variants and policy models. Rows with \textsc{"+ Claude-Sonnet-4"} use \textsc{Claude-Sonnet-4} for the PRM.}
\label{tab:full-results-table}
\resizebox{\columnwidth}{!}{%
\begin{tabular}{@{}llrrrrrrrrrrrrrrrr@{}}
\toprule
\textbf{Setting} &
  \textbf{Model} &
  \multicolumn{1}{l}{\textbf{\begin{tabular}[c]{@{}l@{}}Resolution\\ Rate (\%)\end{tabular}}} &
  \multicolumn{1}{l}{\textbf{\begin{tabular}[c]{@{}l@{}}Easy\\ Resolution\\ Rate (\%)\end{tabular}}} &
  \multicolumn{1}{l}{\textbf{\begin{tabular}[c]{@{}l@{}}Medium\\ Resolution\\ Rate (\%)\end{tabular}}} &
  \multicolumn{1}{l}{\textbf{\begin{tabular}[c]{@{}l@{}}Hard\\ Resolution\\ Rate (\%)\end{tabular}}} &
  \multicolumn{1}{l}{\textbf{\begin{tabular}[c]{@{}l@{}}Patch\\ Generation\\ Rate (\%)\end{tabular}}} &
  \multicolumn{1}{l}{\textbf{\begin{tabular}[c]{@{}l@{}}Avg\\ Steps\end{tabular}}} &
  \multicolumn{1}{l}{\textbf{\begin{tabular}[c]{@{}l@{}}Avg\\ I/P\\ Tokens\end{tabular}}} &
  \multicolumn{1}{l}{\textbf{\begin{tabular}[c]{@{}l@{}}Avg\\ O/P\\ Tokens\end{tabular}}} &
  \multicolumn{1}{l}{\textbf{\begin{tabular}[c]{@{}l@{}}Avg Sup.\\ Invocations\end{tabular}}} &
  \multicolumn{1}{l}{\textbf{\begin{tabular}[c]{@{}l@{}}Avg Sup.\\ I/P Tokens\end{tabular}}} &
  \multicolumn{1}{l}{\textbf{\begin{tabular}[c]{@{}l@{}}Avg Sup.\\ O/P Tokens\end{tabular}}} &
  \multicolumn{1}{l}{\textbf{\begin{tabular}[c]{@{}l@{}}Avg\\ Optimal\\ Windows\end{tabular}}} &
  \multicolumn{1}{l}{\textbf{\begin{tabular}[c]{@{}l@{}}Avg\\ Suboptimal\\ Windows\end{tabular}}} &
  \multicolumn{1}{l}{\textbf{\begin{tabular}[c]{@{}l@{}}Policy\\ Model\\ Cost (\$)\\per 100 \\instances\end{tabular}}} &
  \multicolumn{1}{l}{\textbf{\begin{tabular}[c]{@{}l@{}}Sup.\\ Cost (\$)\\per 100 \\instances\end{tabular}}} &
  \multicolumn{1}{l}{\textbf{\begin{tabular}[c]{@{}l@{}}Total\\Cost (\$)\\per 100 \\instances\end{tabular}}} \\ \midrule

\multirow{4}{*}{\texttt{base}}  &
  \textsc{SWE-agent-LM-32B} &
  40.0 &
  57.2 &
  32.6 &
  8.9 &
  92.4 &
  38.64 &
  340555 &
  5744 &
  - &
  - &
  - &
  - &
  - &
  2.77 &
  - &
  2.77 \\ &
  \textsc{Devstral-Small-2505} &
  34.0 &
  51.0 &
  26.4 &
  4.4 &
  92.6 &
  37.97 &
  330892 &
  5439 &
  - &
  - &
  - &
  - &
  - &
  2.69 &
  - &
  2.69 \\ &
  \textsc{Devstral-Small-2507} &
  30.0 &
  47.4 &
  21.5 &
  4.4 &
  88.0 &
  40.16 &
  332407 &
  5374 &
  - &
  - &
  - &
  - &
  - &
  2.70 &
  - &
  2.70 \\ &
  \textsc{Claude-Sonnet-4} &
  66.6 &
  80.9 &
  62.8 &
  26.7 &
  100.0 &
  61.72 &
  37786 &
  2534 &
  - &
  - &
  - &
  - &
  - &
  121.66 &
  - &
  121.66 \\  \midrule
\multirow{4}{*}{\texttt{SWE-PRM$_{S}$}} &
  \textsc{SWE-agent-LM-32B} &
  19.6 &
  30.4 &
  14.2 &
  4.4 &
  67.6 &
  21.31 &
  254892 &
  2718 &
  4.12 &
  29990 &
  19589 &
  - &
  - &
  2.06 &
  0.40 &
  2.46 \\ &
  \textsc{Devstral-Small-2505} &
  34.4 &
  53.6 &
  24.9 &
  6.7 &
  94.9 &
  41.28 &
  536399 &
  6723 &
  7.92 &
  51627 &
  5023 &
  - &
  - &
  4.34 &
  0.45 &
  4.80 \\ &
  \textsc{Devstral-Small-2507} &
  33.6 &
  50.5 &
  25.3 &
  8.9 &
  93.4 &
  45.54 &
  544035 &
  6492 &
  8.69 &
  49523 &
  4651 &
  - &
  - &
  4.40 &
  0.43 &
  4.84 \\ &
  \multicolumn{1}{l}{\begin{tabular}[c]{@{}l@{}}\textsc{SWE-agent-LM-32B} \\$+$ \textsc{Claude-Sonnet-4}\end{tabular}} &
  45.8 &
  63.4 &
  39.1 &
  8.9 &
  98.2 &
  51.54 &
  593077 &
  7419 &
  10.0 &
  60192 &
  3706 &
  - &
  - &
  4.80 &
  23.62 &
  28.42 \\ \midrule
\multirow{3}{*}{\texttt{SWE-PRM$_{C}$}} &
  \textsc{SWE-agent-LM-32B} &
  35.6 &
  54.1 &
  26.8 &
  6.7 &
  91.4 &
  34.32 &
  419819 &
  4674 &
  6.49 &
  41894 &
  5084 &
  0.37 &
  6.13 &
  3.40 &
  0.38 &
  3.77 \\ &
  \textsc{Devstral-Small-2505} &
  34.2 &
  54.1 &
  24.9 &
  2.2 &
  92.2 &
  38.39 &
  438097 &
  5326 &
  7.34 &
  47719 &
  3801 &
  0.84 &
  6.50 &
  3.55 &
  0.41 &
  3.96 \\ &
  \textsc{Devstral-Small-2507} &
  30.2 &
  47.9 &
  21.5 &
  4.4 &
  90.2 &
  43.46 &
  498381 &
  6551 &
  8.30 &
  48540 &
  3815 &
  0.34 &
  7.96 &
  4.04 &
  0.42 &
  4.46 \\ \midrule
\multirow{3}{*}{\texttt{SWE-PRM$_{CG}$}} &
  \textsc{SWE-agent-LM-32B} &
  35.6 &
  52.1 &
  28.7 &
  4.4 &
  89.8 &
  32.71 &
  344833 &
  4426 &
  6.19 &
  40824 &
  5274 &
  0.64 &
  5.55 &
  2.79 &
  0.37 &
  3.16 \\ &
  \textsc{Devstral-Small-2505} &
  34.2 &
  54.1 &
  24.9 &
  2.2 &
  92.8 &
  37.65 &
  354389 &
  5106 &
  7.19 &
  45723 &
  3743 &
  0.95 &
  6.24 &
  2.88 &
  0.40 &
  3.27 \\ &
  \textsc{Devstral-Small-2507} &
  30.2 &
  47.9 &
  21.5 &
  4.4 &
  91.0 &
  41.52 &
  409703 &
  5887 &
  7.88 &
  46991 &
  3633 &
  0.53 &
  7.36 &
  3.32 &
  0.40 &
  3.73 \\ \midrule
\multirow{4}{*}{\texttt{SWE-PRM$_{D}$}} &
  \textsc{SWE-agent-LM-32B} &
  38.8 &
  56.2 &
  31.4 &
  6.7 &
  92.2 &
  33.12 &
  360688 &
  4510 &
  6.18 &
  44751 &
  3262 &
  0.42 &
  5.77 &
  2.92 &
  0.38 &
  3.31 \\ &
  \textsc{Devstral-Small-2505} &
  34.2 &
  54.1 &
  24.9 &
  2.2 &
  93.4 &
  37.89 &
  421554 &
  5752 &
  7.22 &
  52587 &
  2826 &
  0.37 &
  6.85 &
  3.42 &
  0.44 &
  3.86 \\ &
  \textsc{Devstral-Small-2507} &
  30.2 &
  47.9 &
  21.5 &
  4.4 &
  93.4 &
  40.08 &
  457684 &
  6338 &
  7.63 &
  51242 &
  3391 &
  0.31 &
  7.32 &
  3.71 &
  0.44 &
  4.15 \\ &
  
  \multicolumn{1}{l}{\begin{tabular}[c]{@{}l@{}}\textsc{SWE-agent-LM-32B} \\$+$ \textsc{Claude-Sonnet-4}\end{tabular}}&
  50.6 &
  69.1 &
  43.3 &
  13.3 &
  98.2 &
  37.99 &
  446185 &
  5674 &
  7.24 &
  51443 &
  4621 &
  0.03 &
  7.21 &
  3.61 &
  2.37 &
  25.98 \\ \midrule
\multirow{3}{*}{\texttt{SWE-PRM$_{DN}$}} &
  \textsc{SWE-agent-LM-32B} &
  30.0 &
  41.8 &
  25.7 &
  4.4 &
  79.6 &
  27.54 &
  350412 &
  3407 &
  5.21 &
  36686 &
  7306 &
  0.47 &
  4.73 &
  2.83 &
  0.35 &
  3.18 \\ &
  \textsc{Devstral-Small-2505} &
  34.2 &
  54.1 &
  24.9 &
  2.2 &
  94.4 &
  37.72 &
  450821 &
  5408 &
  7.13 &
  47078 &
  4665 &
  0.65 &
  6.48 &
  3.65 &
  0.41 &
  4.06 \\ &
  \textsc{Devstral-Small-2507} &
  30.2 &
  47.9 &
  21.5 &
  4.4 &
  91.6 &
  39.98 &
  505866 &
  6266 &
  7.63 &
  48816 &
  5092 &
  0.69 &
  6.93 &
  4.10 &
  0.43 &
  4.53 \\ \midrule
\multirow{3}{*}{\texttt{SWE-PRM$_{DG}$}} &
  \textsc{SWE-agent-LM-32B} &
  34.8 &
  51.5 &
  28 &
  2.2 &
  93.2 &
  33.82 &
  325223 &
  4519 &
  5.65 &
  39090 &
  2793 &
  0.88 &
  4.77 &
  2.64 &
  0.34 &
  2.97 \\ &
  \textsc{Devstral-Small-2505} &
  34.2 &
  54.1 &
  24.9 &
  2.2 &
  95.4 &
  38.58 &
  371001 &
  5405 &
  7.39 &
  54926 &
  2854 &
  1.00 &
  6.39 &
  3.01 &
  0.46 &
  3.47 \\ &
  \textsc{Devstral-Small-2507} &
  30.2 &
  47.9 &
  21.5 &
  4.4 &
  93 &
  39.52 &
  364568 &
  5557 &
  7.50 &
  50056 &
  3325 &
  1.00 &
  6.50 &
  2.96 &
  0.43 &
  3.39 \\ \midrule
\multirow{3}{*}{\texttt{SWE-PRM$_{DNG}$}} &
  \textsc{SWE-agent-LM-32B} &
  30.0 &
  41.8 &
  25.7 &
  4.4 &
  54.8 &
  10.11 &
  110855 &
  1118 &
  1.93 &
  19379 &
  21792 &
  0.78 &
  1.15 &
  0.90 &
  0.33 &
  1.23 \\ &
  \textsc{Devstral-Small-2505} &
  34.2 &
  54.1 &
  24.9 &
  2.2 &
  94.4 &
  36.05 &
  354803 &
  5229 &
  6.80 &
  46335 &
  4412 &
  0.99 &
  5.80 &
  2.88 &
  0.41 &
  3.29 \\ &
  \textsc{Devstral-Small-2507} &
  30.4 &
  47.9 &
  21.8 &
  4.4 &
  91.8 &
  39.22 &
  365504 &
  5260 &
  7.44 &
  46252 &
  5066 &
  0.98 &
  6.46 &
  2.97 &
  0.41 &
  3.38 \\ \midrule
\multirow{4}{*}{\texttt{SWE-PRM$_{DR}$}} &
  \textsc{SWE-agent-LM-32B} &
  36.8 &
  50.5 &
  31.8 &
  6.7 &
  92.8 &
  28.67 &
  299191 &
  3900 &
  5.44 &
  44223 &
  4767 &
  0.49 &
  4.95 &
  2.42 &
  0.39 &
  2.82 \\ &
  \textsc{Devstral-Small-2505} &
  36.0 &
  51.5 &
  30.3 &
  2.2 &
  95.0 &
  32.33 &
  326033 &
  4300 &
  6.17 &
  49445 &
  2287 &
  0.45 &
  5.72 &
  2.64 &
  0.41 &
  3.06 \\ &
  \textsc{Devstral-Small-2507} &
  32.4 &
  51.0 &
  23.4 &
  4.4 &
  94.4 &
  37.67 &
  418660 &
  5148 &
  7.14 &
  56343 &
  3187 &
  0.37 &
  6.78 &
  3.39 &
  0.48 &
  3.87 \\ &
  \multicolumn{1}{l}{\begin{tabular}[c]{@{}l@{}}\textsc{SWE-agent-LM-32B} \\$+$ \textsc{Claude-Sonnet-4}\end{tabular}} &
  44.8 &
  62.9 &
  36.8 &
  13.3 &
  98.2 &
  34.38 &
  389420 &
  4984 &
  6.39 &
  52193 &
  3810 &
  0.02 &
  6.37 &
  3.16 &
  21.37 &
  24.53 \\ \bottomrule
\end{tabular}%
}
\end{table}
\end{landscape}
}

\end{document}